\documentclass[letterpaper]{article} 
\usepackage{aaai2026}  
\usepackage{times}  
\usepackage{helvet}  
\usepackage{courier}  
\usepackage[hyphens]{url}  
\usepackage{multirow} 
\usepackage{graphicx} 
\usepackage{booktabs}
\usepackage{amsmath}
\usepackage{float}
\usepackage{natbib}  
\usepackage{caption} 
\usepackage{algorithm}
\usepackage{algorithmic}
\usepackage{enumitem}

\usepackage{newfloat}
\usepackage{listings}
\DeclareCaptionStyle{ruled}{labelfont=normalfont,labelsep=colon,strut=off} 
\floatstyle{ruled}
\newfloat{listing}{tb}{lst}{}
\floatname{listing}{Listing}
\title{Evaluating Autoformalization Robustness via Semantically \\Similar Paraphrasing}
\author{
    Hayden Moore\textsuperscript{\rm 1}\equalcontrib,
    Asfahan Shah\textsuperscript{\rm 1}\equalcontrib
}
\affiliations{
    \textsuperscript{\rm 1}Department of Computer Science and Engineering, The Pennsylvania State University \\
    University Park, PA 16802 USA\\
    \{hmm5731, aks7824\}@psu.edu

}

\usepackage{bibentry}

\begin{document}

\maketitle

\begin{abstract}
Large Language Models (LLMs) have recently emerged as powerful tools for autoformalization. Despite their impressive performance, these models can still struggle to produce grounded and verifiable formalizations. Recent work in text-to-SQL, has revealed that LLMs can be sensitive to paraphrased natural language (NL) inputs, even when high degrees of semantic fidelity are preserved.
In this paper, we investigate this claim in the autoformalization domain. Specifically, we evaluate the robustness of LLMs generating formal proofs with semantically similar paraphrased NL statements by measuring semantic and compilation validity. Using the formal benchmarks MiniF2F and Lean 4 version of ProofNet, and two modern LLMs, we generate paraphrased natural language statements and cross-evaluate these statements across both models. The results of this paper reveal performance variability across paraphrased inputs, demonstrating that minor shifts in NL statements can significantly impact model outputs.
\end{abstract}

\section{Introduction}
Recent research has shown progress in \textit{autoformalization}, defined as the translation of natural language (NL) mathematical statements into formal proofs. By combining modern large language models (LLMs) with symbolic theorem provers such as Isabelle \cite{isabelle} and Lean \cite{lean}, there is hope for making formal mathematical verifications accessible to anyone who can articulate a problem in NL. Closing the gap between a novice understanding of math and formal proofs, changing how knowledge could be authored and verified.

Despite these recent advances, using LLMs for autoformalization remains far from a verifiable and trusted system that is deployable in practice. Currently, these systems often generate formal statements that could be syntactically valid but remain unverifiable, logically inconsistent, or completely wrong. Things like unintended insertions, omissions, misinterpretations, or hallucinations demonstrate not only errors in translation but deeper failures of foundational model integrity. This is especially important in a domain like mathematics, where correctness is an absolute requirement, and where one small error can undermine the reliability of the entire generated proof chain.

These errors can erode trust and limit adoption into real-world applications. To adopt LLM autoformalization into a serious workflow, we must first validate that these systems produce correct outputs and be able to explain why its correct. Without this necessary interpretability and explainability, the future of machine assisted formal reasoning risks becoming a black-box system, which experts typically hesitate to adopt. Only by embedding explainability and interpretability at the core of these systems and benchmarks can we help ensure that the future of autoformalization enhances without obscuring our understanding of the system.

\begin{figure*}[t]
  \centering
  \includegraphics[width=\textwidth]{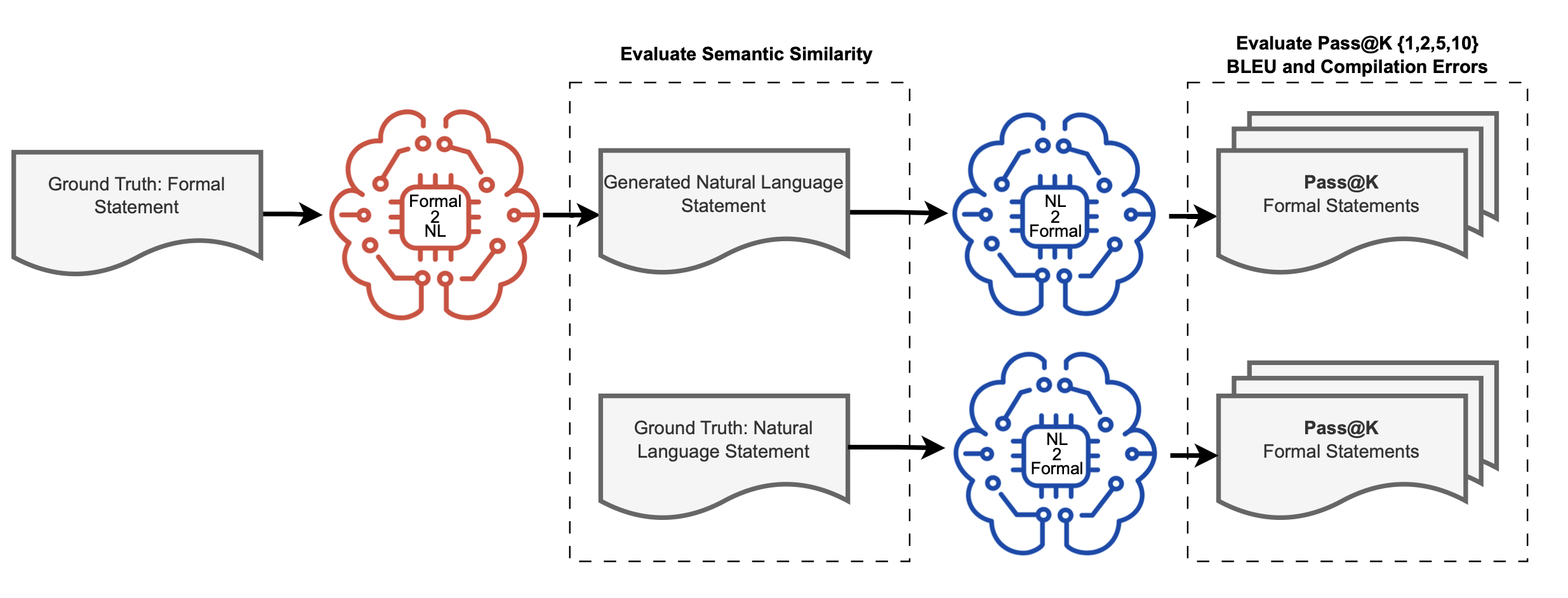}
  \caption{Overview of the autoformalization robustness evaluation pipeline for MiniF2F(Isabelle/HOL) and ProofNet(Lean 4). Each formal system undergoes both forward (Formal→NL) and reverse (NL→Formal) paraphrasing stages, using both GPT-4o-mini \cite{gpt} and Claude-3.7-sonnet \cite{claude}, with consistency evaluated via similarity metrics and Pass@K accuracy.}
  \label{fig:overview}
\end{figure*}

\section{Related Work}

Recent research has reported that linguistic variations can lead to significant performance degradation in the text-to-SQL domain, stating \textit{"This problem underscores the sensitivity of current models and the need for more resilient solutions"} \cite{oracle-sql}. This begs the question of whether such sensitivity extends to other domains that also rely on LLMs as their core component.

In the domain of mathematics, the concept of autoformalization using LLMs was first demonstrated by \cite{LLM}, where LLMs were used to formalize natural-language mathematical statements into Isabelle/HOL. Since then, impressive progress has been made in advancing autoformalization techniques. Building on these advances \cite{leandojo} developed LeanDojo, which provides unified infrastructure and datasets for training and evaluating LLMs on formalization tasks in Lean. Similarly \cite{thor} introduced Thor, demonstrating how LLMs can be integrated with automated theorem provers to solve formal proofs in Isabelle.

The MiniF2F benchmark \cite{minif2f} introduced a dataset for Olympiad-level formal reasoning enabling evaluation of LLM performance in proof synthesis for formal languages such as Isabelle \cite{isabelle}. Additionally, ProofNet \cite{proof} expanded the scale and diversity of mathematical problems, bridging undergraduate-level mathematics with formal proofs to evaluate LLM's performance in proof synthesis for Lean 3 code \cite{lean}. \cite{lean4} later introduced a Lean 4 version of ProofNet.

\section{Preliminaries}

\paragraph{BLEU Evaluation.}
We adopt the BLEU formalization used in \citet{minif2f}, this metric was originally proposed by \citet{papineni2002bleu}.
Given a reference sequence $r$ and a candidate sequence $c$, the BLEU score is defined as:
\begin{equation}
\mathrm{BLEU} = \mathrm{BP} \cdot \exp\!\left( \sum_{n=1}^{N} w_n \log p_n \right),
\end{equation}
where $p_n$ denotes the modified $n$-gram precision up to order $N$, $w_n$ is the weight (typically $w_n = \tfrac{1}{N}$),
and $\mathrm{BP}$ is the brevity penalty:
\begin{equation}
\mathrm{BP} =
\begin{cases}
1, & \text{if } c_{\text{len}} > r_{\text{len}}, \\
\exp\!\left( 1 - \tfrac{r_{\text{len}}}{c_{\text{len}}} \right), & \text{otherwise.}
\end{cases}
\end{equation}
We follow the MiniF2F and ProofNet evaluation protocol by computing sentence-level BLEU with smoothing (method 1 from NLTK)
to account for sparse $n$-gram overlap in short formal proofs.

\paragraph{Lexical Diversity.}
To quantify variation in word usage across paraphrased and reference statements, we compute
\textit{lexical diversity} as the type–token ratio (TTR) \cite{lex}:
\begin{equation}
\mathrm{TTR} = \frac{|V|}{|W|},
\end{equation}
where $|V|$ is the number of unique tokens and $|W|$ is the total number of tokens
in the natural language statement.

\paragraph{Cosine Similarity.}
To measure semantic equivalence between the paraphrased and reference statements, we compute
the cosine similarity of their sentence embeddings:
\begin{equation}
\mathrm{Sim}_{\cos}(r, c) = \frac{E(r) \cdot E(c)}{\|E(r)\| \|E(c)\|},
\end{equation}
where $E(\cdot)$ denotes the SBERT \cite{sbert} embedding function.

\section{Methodology}

For our methodology we follow a two staged process to evaluate how well autoformalization performs when introduced with paraphrased inputs (Figure \ref{fig:overview}). \textbf{Each stage is context independent, meaning that all LLM requests do not have context to previous requests.}

The first stage is where we generate the paraphrased NL statements using two modern LLMs, GPT-4o-mini \cite{gpt} and Claude-3.7-sonnet \cite{claude}. Each formal statement from MiniF2F and ProofNet (Lean 4), is passed as input to the LLMs, prompting them to translate the formal proof into a NL statement that accurately captures the logic of the proof (prompts defined in the Appendix). We also perform a semantic similarity analysis on these paraphrased NL statements and the corresponding ground truth NL statement, which is defined in the results section of this paper. This stage establishes the semantic validity of the paraphrased NL statements before we evaluate the translation performance for NL→Formal.

The second stage is where we perform a Pass@K cross-evaluation of GPT and Claude paraphrased NL statements. Testing each models ability and sensitivity to handling variations of the original NL statement that are still semantically similar. We pass GPT/Claude paraphrased NL statements to both GPT/Claude models and evaluate their performance on both BLEU accuracy and compilation accuracy. BLEU accuracy is defined in the previous preliminaries section. Compilation accuracy is defined by the successful execution of the generated proof code (Isabelle/Lean 4) with the respective compilers.

\section{Results}
\subsection{Semantic Similarity Analysis}

\begin{figure}[H]
  \centering
  \setlength{\tabcolsep}{2pt}%
  \begin{tabular}{@{}cc@{}}
    \includegraphics[width=0.48\columnwidth]{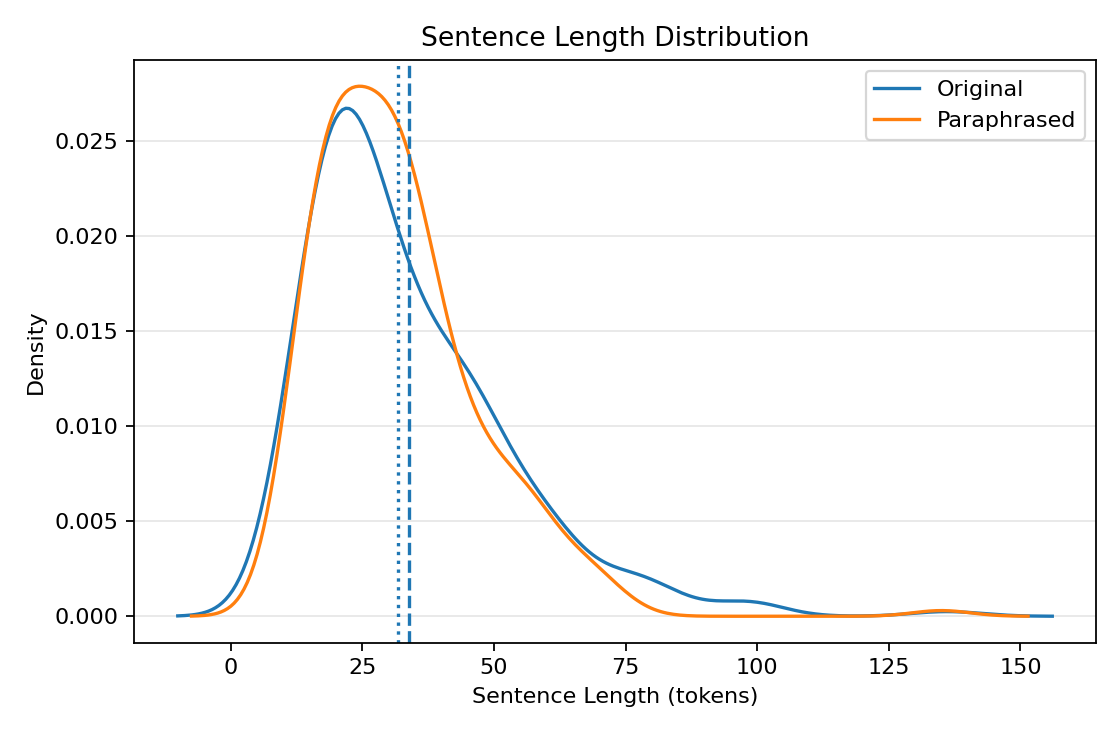} &
    \includegraphics[width=0.48\columnwidth]{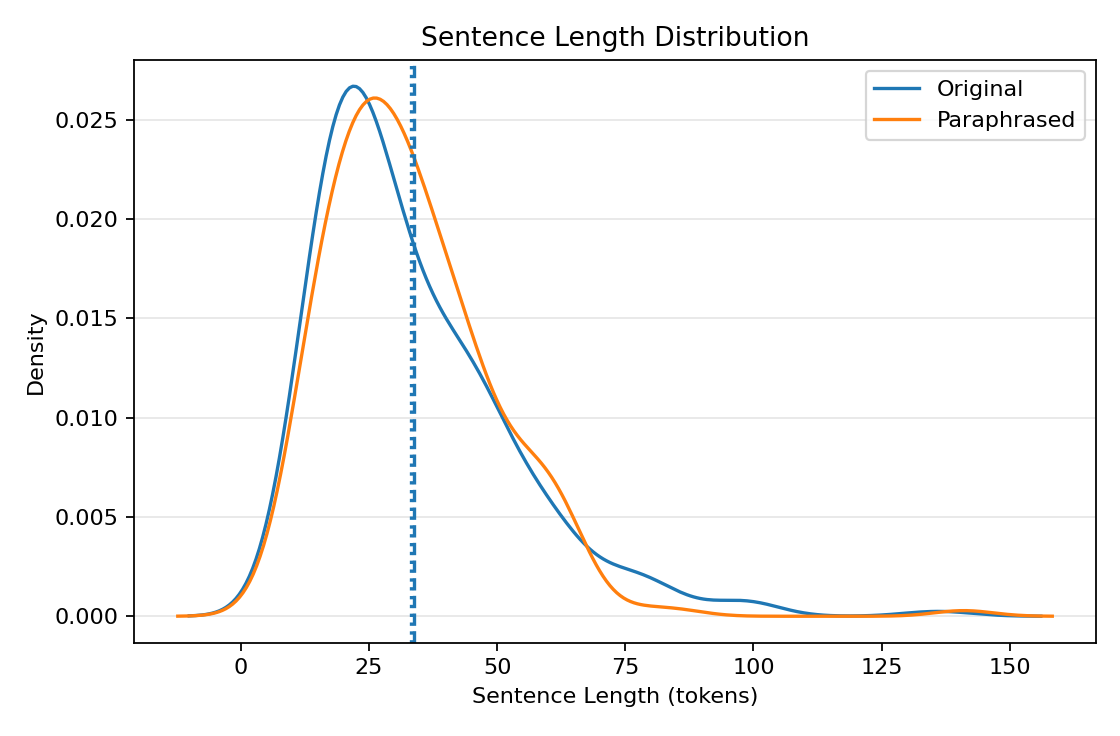} \\
    \footnotesize (a) GPT-4o-mini&
    \footnotesize (b) Claude-3.7-sonnet\\
    \includegraphics[width=0.48\columnwidth]{\detokenize{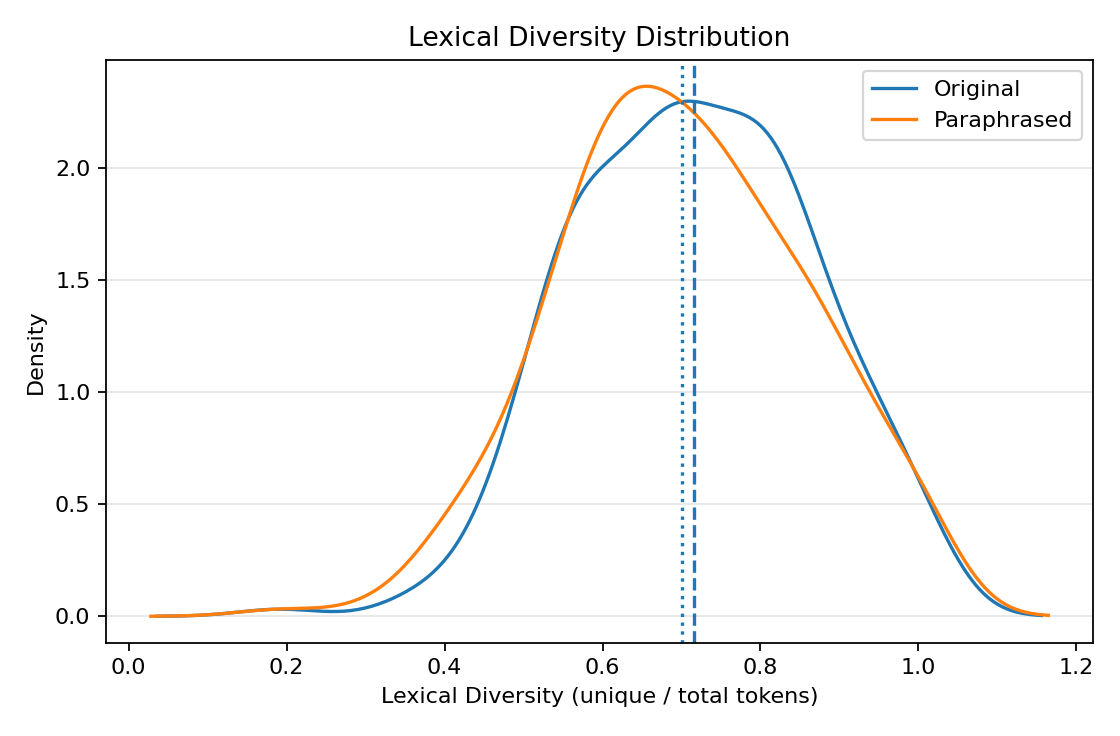}} &
    \includegraphics[width=0.48\columnwidth]{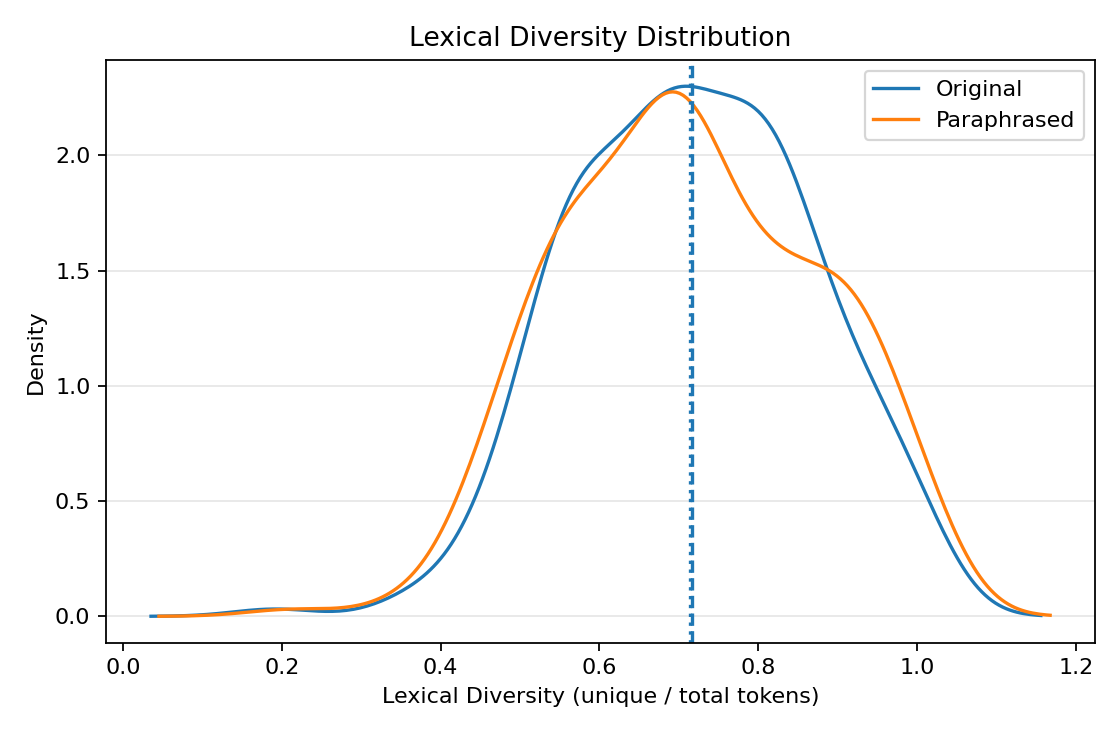} \\
    \footnotesize (c) GPT-4o-mini&
    \footnotesize (d) Claude-3.7-sonnet\\
    \includegraphics[width=0.48\columnwidth]{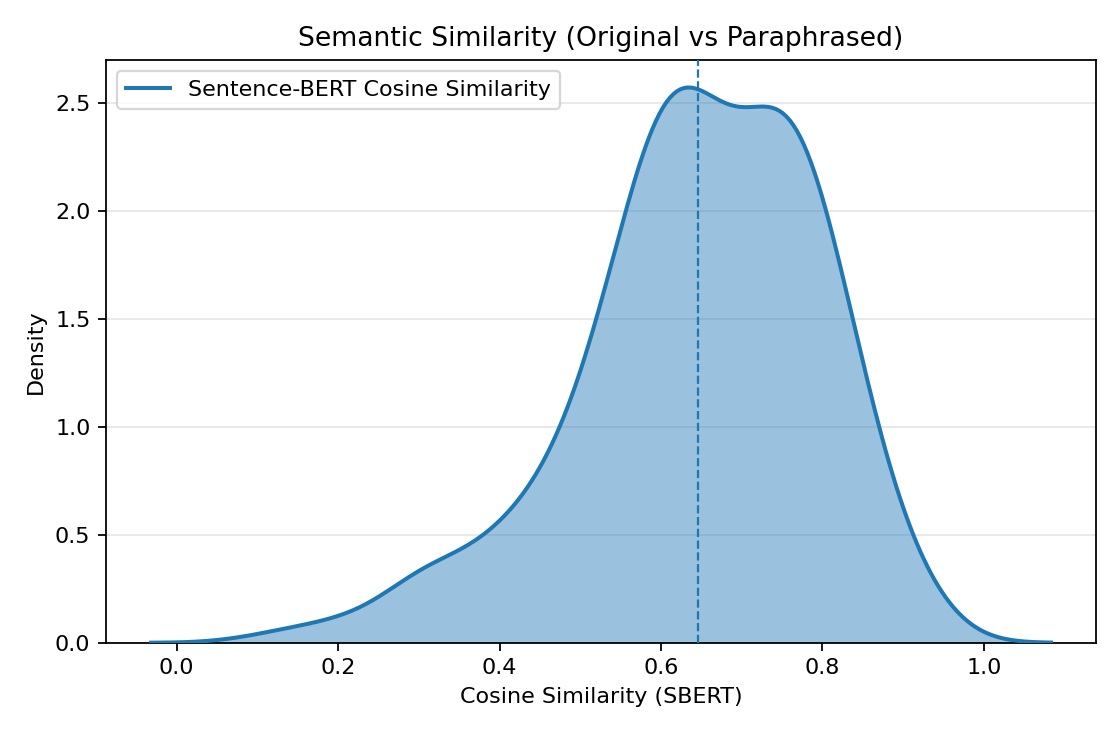} &
    \includegraphics[width=0.48\columnwidth]{\detokenize{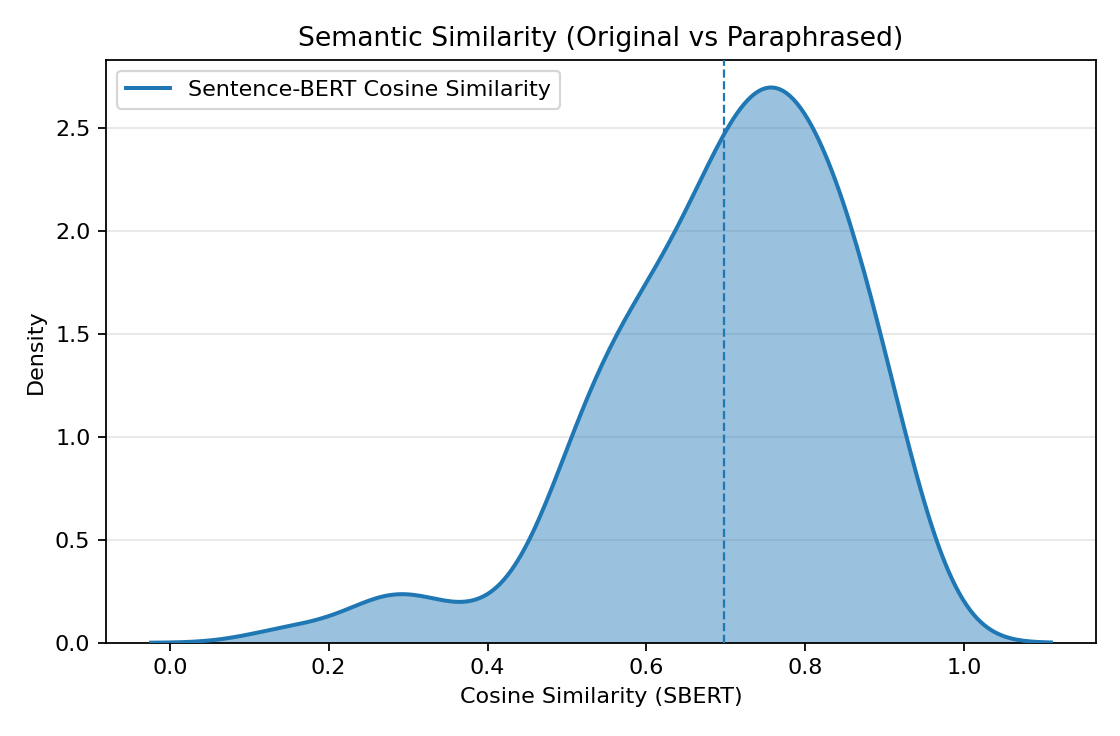}} \\
    \footnotesize (e) GPT-4o-mini&
    \footnotesize (f) Claude-3.7-sonnet\\
  \end{tabular}
  \caption{\textbf{MiniF2F (Isabelle)}:
  Panels show sentence length, lexical diversity, and SBERT semantic similarity for GPT-4o-mini and Claude-3.7 paraphrasings.}
  \label{fig:corpus_comparison_column}
\end{figure}
We first create a baseline of the semantic similarity between the paraphrased NL statements and their corresponding ground truth formal NL statements (Figure \ref{fig:corpus_comparison_column}). 

Our results with \textbf{MiniF2F (Isabelle)}  using both GPT-4o-mini and Claude-3.7-sonnet for paraphrasing, we observe consistent and high cosine similarity distributions relative to the ground truth NL statements (64-72\%). Indicating that the paraphrasing step largely preserves semantic meaning even across the diversity of the benchmark. 

\begin{figure}[H]
  \centering
  \setlength{\tabcolsep}{2pt}%
  \begin{tabular}{@{}cc@{}}
    \includegraphics[width=0.48\columnwidth]{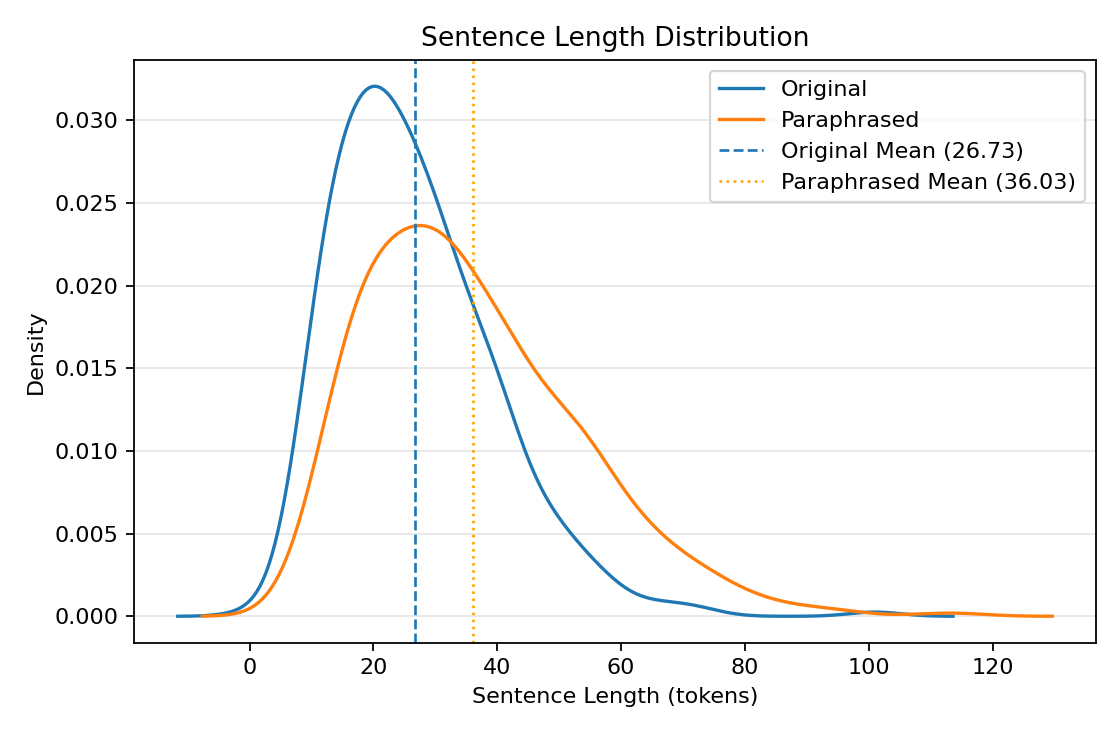} &
    \includegraphics[width=0.48\columnwidth]{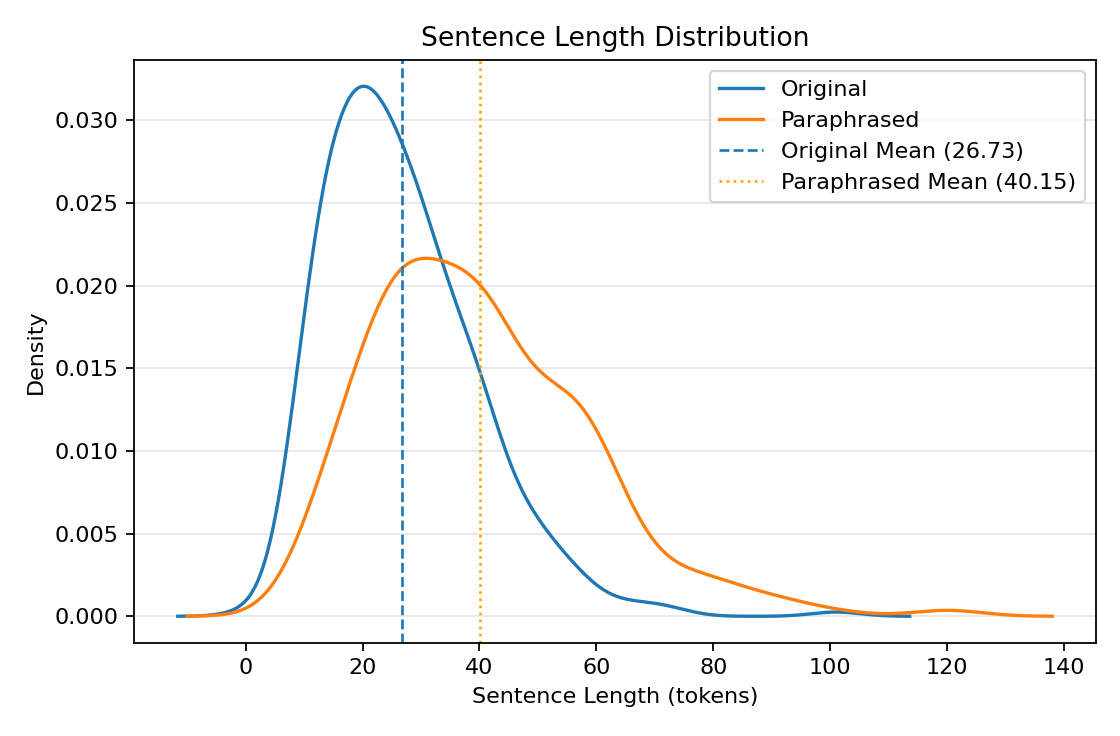} \\
    \footnotesize (a) GPT-4o-mini&
    \footnotesize (b) Claude-3.7-sonnet\\
    \includegraphics[width=0.48\columnwidth]{\detokenize{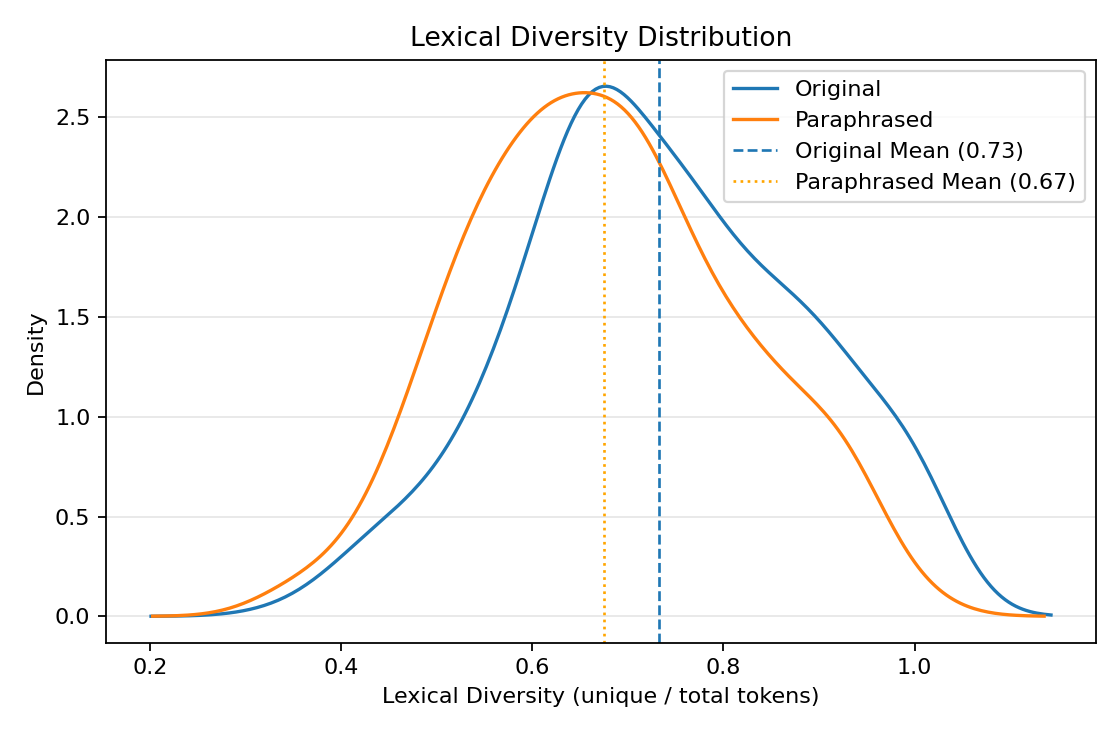}} &
    \includegraphics[width=0.48\columnwidth]{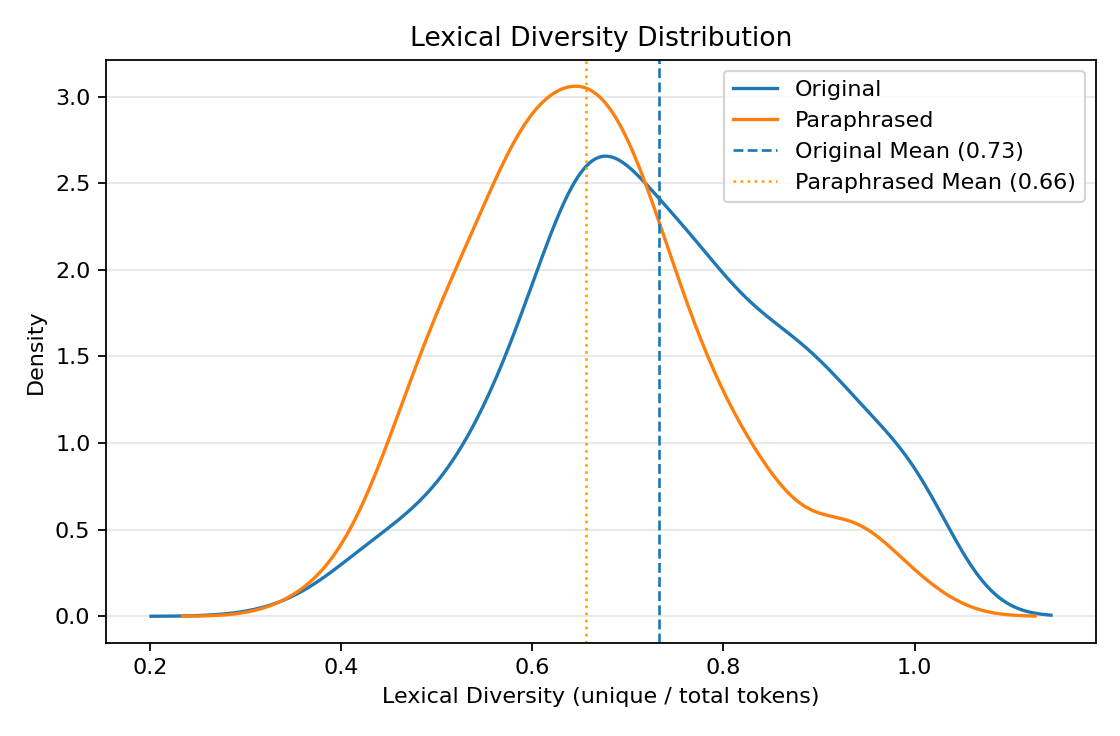} \\
    \footnotesize (c) GPT-4o-mini&
    \footnotesize (d) Claude-3.7-sonnet\\
    \includegraphics[width=0.48\columnwidth]{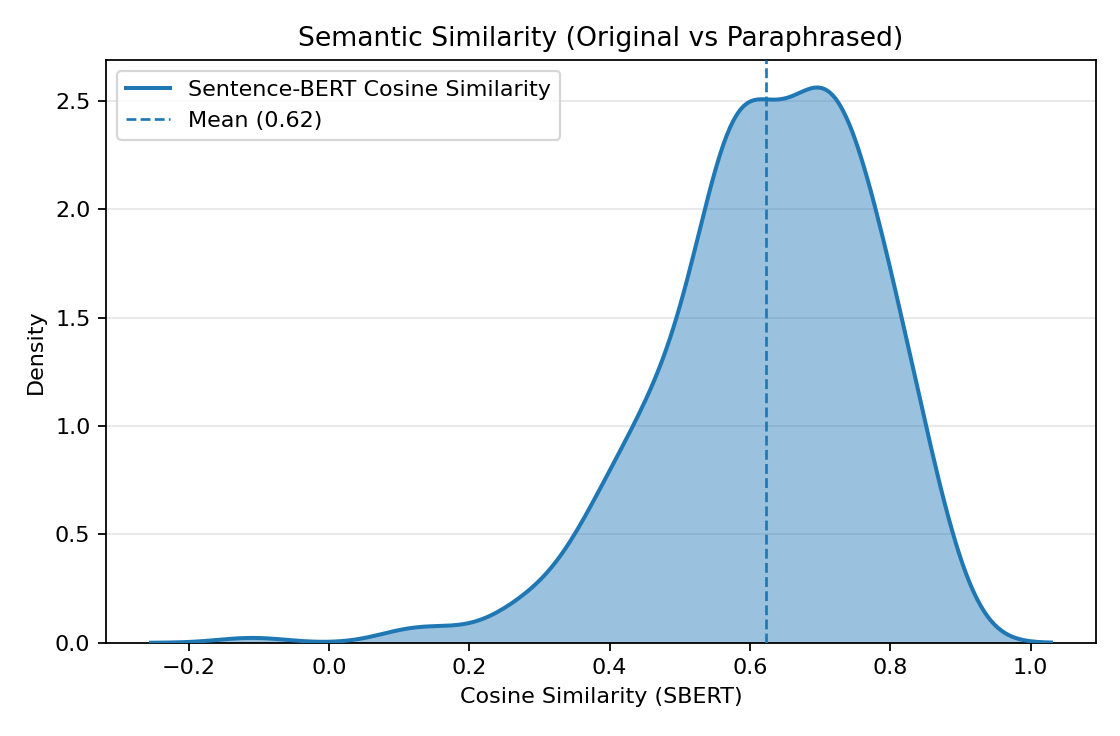} &
    \includegraphics[width=0.48\columnwidth]{\detokenize{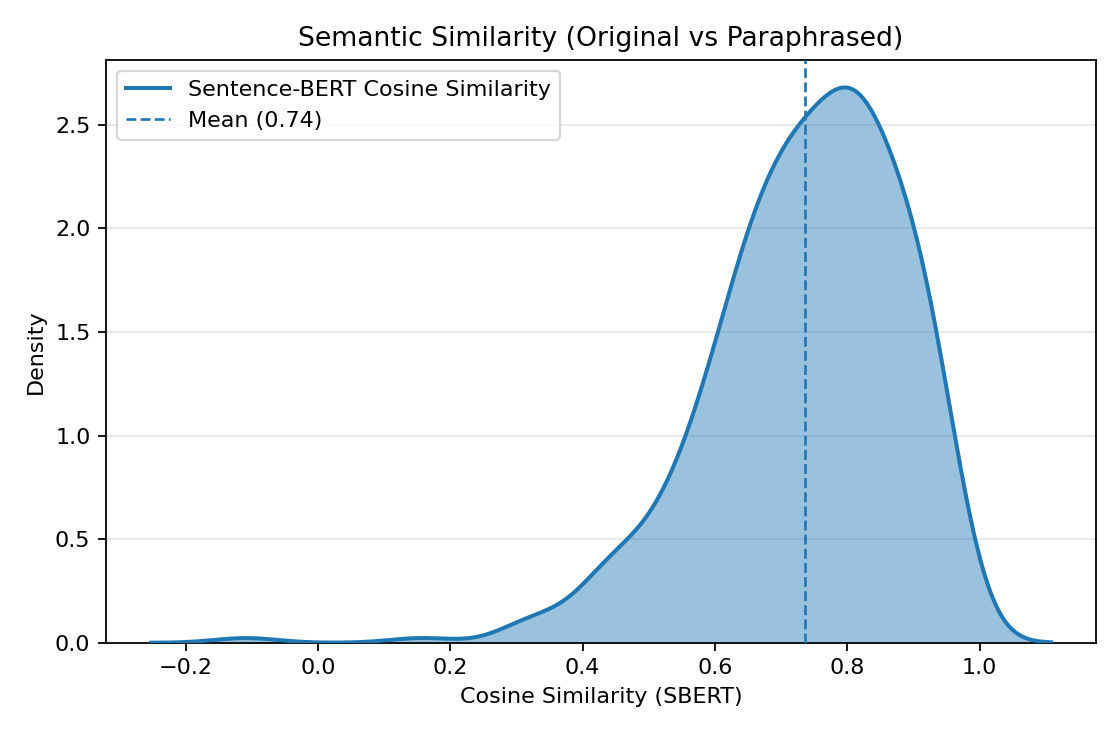}} \\
    \footnotesize (e) GPT-4o-mini&
    \footnotesize (f) Claude-3.7-sonnet\\
  \end{tabular}
  \caption{\textbf{ProofNet (Lean 4)}: Panels show sentence length, lexical diversity, and SBERT semantic similarity for GPT-4o-mini and Claude-3.7 paraphrasings.}
  \label{fig:corpus_comparison_column_lean}
\end{figure}

For \textbf{ProofNet (Lean 4)} (Figure \ref{fig:corpus_comparison_column_lean}), when using both GPT-4o-mini and Claude-3.7-sonnet for paraphrasing, we also observe high cosine similarity distributions relative to the ground truth NL statements (62-78\%). Indicating once again that the paraphrasing step preserves semantic meaning. However, we do observe a more loose clustering around the centroid for lexical diversity and sentence length, indicating that the Formal→NL translation for Lean 4 seems to introduce more lexical diversity.

These results help us confirm that the paraphrasing stage maintains a high semantic equivalence across both models tested while still providing linguistic variations to the statement. This helps ensure that our subsequent variations in formalizations stems less from semantic meaning drift but more from the LLMs sensitivity to the linguistic differences in the paraphrased statements. In other words, the LLMs are given semantically equivalent paraphrased statements and we test if the autoformalization step remains invariant.

\subsection{Pass@K Evaluation}
\begin{figure}[H]
  \centering
  \setlength{\tabcolsep}{2pt}%
  \begin{tabular}{@{}cc@{}}
    \includegraphics[width=0.49\columnwidth]{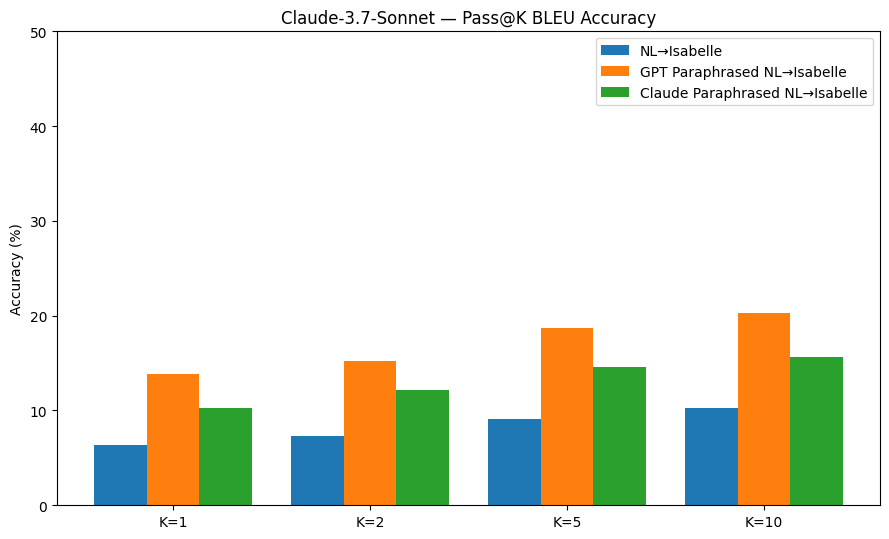} &
    \includegraphics[width=0.49\columnwidth]{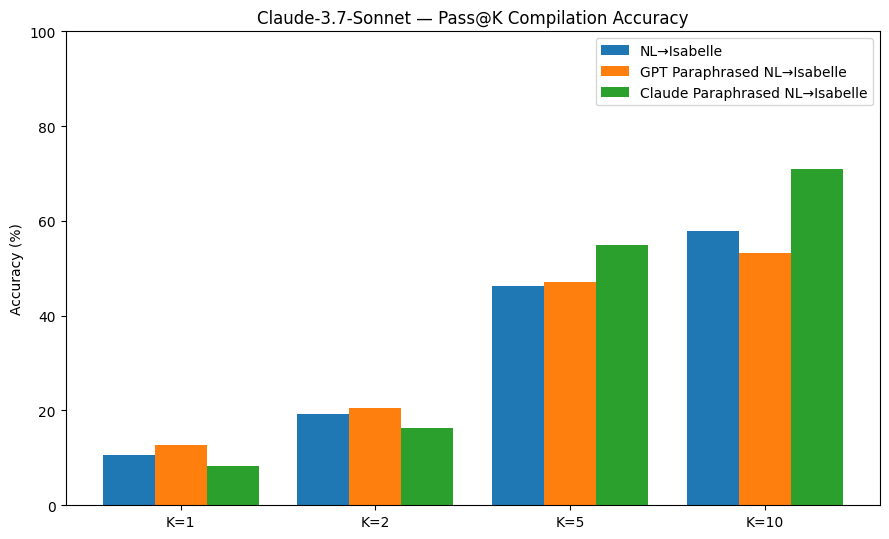} \\
    \footnotesize (a) Claude-3.7-Sonnet, BLEU&
    \footnotesize (b) Claude-3.7-Sonnet, Comp.\\
    \includegraphics[width=0.49\columnwidth]{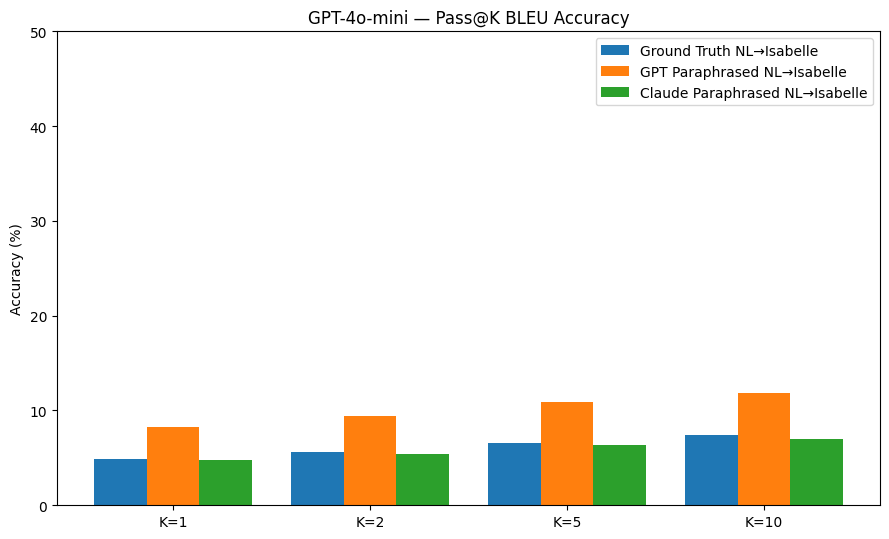} &
    \includegraphics[width=0.49\columnwidth]{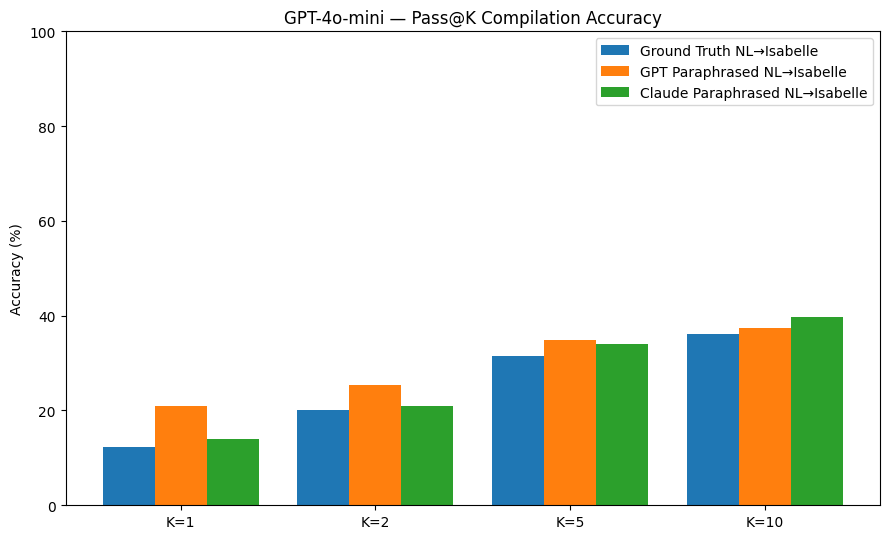} \\
    \footnotesize (c) GPT-4o-mini, BLEU &
    \footnotesize (d) GPT-4o-mini, Comp.\\
  \end{tabular}
  \caption{\textbf{MiniF2F (Isabelle)} autoformalization cross-evaluation results.
  Pass@K accuracy is reported for $K\!\in\!\{1,2,5,10\}$ across semantic (BLEU) and syntactic (Compilation) metrics.
  Results are shown for both formalization models: Claude-3.7-Sonnet and GPT-4o-mini.}
  \label{fig:passk_grid}
\end{figure}

The Pass@K results in Table \ref{tab:passk_combined} reveal clear cross-model dependencies between the paraphraser and the formalization model. While both models exhibit improved average BLEU and compilation performance when using paraphrased inputs, the magnitude and direction of these improvements appear to be sensitive to the paraphrase source.
\paragraph{MiniF2F: Isabelle/HOL} Our results from the MiniF2F benchmark experiments reveal clear variability in performance across paraphrased inputs (Figure \ref{fig:passk_grid}). First, when GPT-4o-mini is used as the formalization model, GPT-paraphrased inputs lead to higher BLEU for all $K$ and compilation scores for Pass@\{1,2,5\} compared to the ground-truth and Claude-paraphrased inputs. When the GPT paraphrases are evaluated by Claude-3.7-sonnet, the semantic fidelity (BLEU) remains strong, while compilation accuracy is stronger at Pass@\{1,2\} but less pronounced at Pass@\{5,10\} and having the lowest Pass@10 score for the Claude formalization experiments. However, if we just look at the BLEU scores for the GPT-paraphrased NL statements, we see across the board, for all $K$, they are the highest.

Conversely, Claude-paraphrased inputs exhibit stronger generalization effects for compilation accuracy. The model achieves 70.9\% Pass@10 compilation, which is the highest score across all configurations. Interestingly, BLEU scores for Claude paraphrases lag slightly behind GPT paraphrases, implying that while Claude’s phrasing improves logical structure and compilation validity, it sometimes departs further from the distributions of the reference statement. When evaluated by GPT-4o-mini, these same paraphrases retain moderate BLEU similarity and display good compilation accuracy at Pass@\{5,10\}.

\begin{figure}[H]
  \centering
  \setlength{\tabcolsep}{2pt}%
  \begin{tabular}{@{}cc@{}}
    \includegraphics[width=0.49\columnwidth]{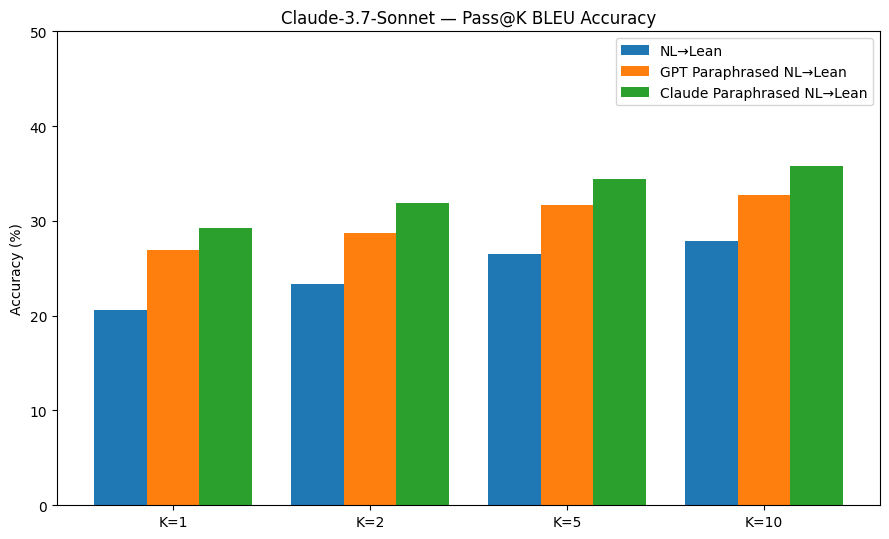} &
    \includegraphics[width=0.49\columnwidth]{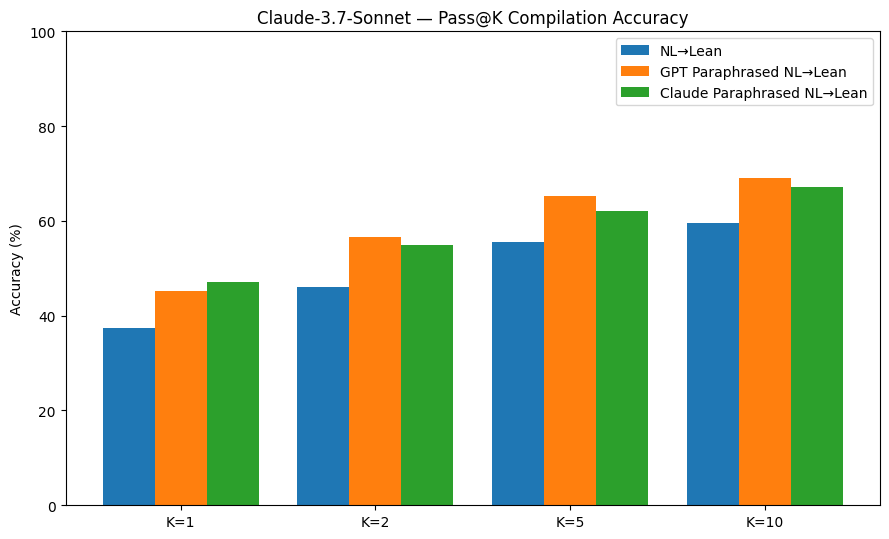} \\
    \footnotesize (a) Claude-3.7-Sonnet, BLEU&
    \footnotesize (b) Claude-3.7-Sonnet, Comp.\\
    \includegraphics[width=0.49\columnwidth]{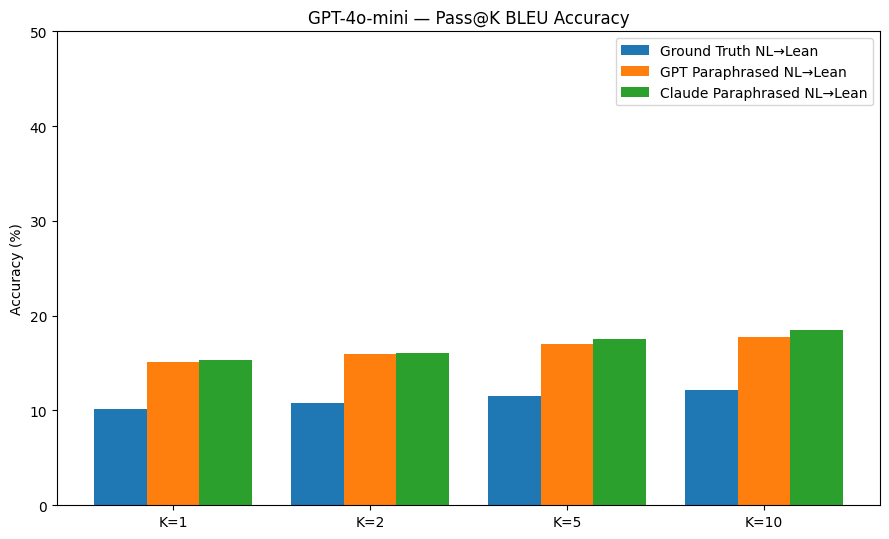} &
    \includegraphics[width=0.49\columnwidth]{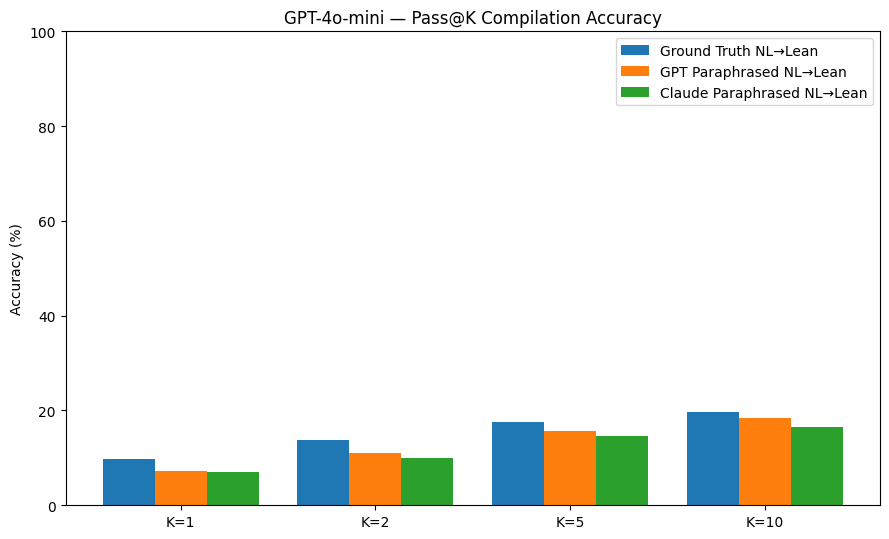} \\
    \footnotesize (c) GPT-4o-mini, BLEU &
    \footnotesize (d) GPT-4o-mini, Comp.\\
  \end{tabular}
  \caption{\textbf{ProofNet (Lean 4)} autoformalization cross-evaluation results.
  Pass@K accuracy is reported for $K\!\in\!\{1,2,5,10\}$ across semantic (BLEU) and syntactic (Compilation) metrics.
  Results are shown for both formalization models: Claude-3.7-Sonnet and GPT-4o-mini.}
  \label{fig:passk_grid_lean}
\end{figure}

\paragraph{ProofNet: Lean 4}
Our results from the ProofNet (Lean 4) benchmark experiments also reveal variability in performance across paraphrased inputs (Figure \ref{fig:passk_grid_lean}). First, the magnitude of both BLEU and compilation changes is more pronounced than with our Isabelle experiments. When GPT-4o-mini is the formalization model, both paraphrased inputs improve BLEU over the ground-truth NL statements across all $K$. However, compilation accuracy for GPT-4o-mini declines slightly under paraphrasing, indicating that subtle syntactic shifts may have been introduced that affect executability in Lean 4’s strict type system. These findings highlight that paraphrasing can sometimes succeed with meaning preservation but does not always lead to syntactic validity.

In contrast, Claude-3.7-Sonnet demonstrates slightly stronger robustness and transferability across paraphrased inputs. BLEU steadily increases from 20.58\% (ground-truth) to 29.25\% under Claude paraphrasing, and compilation accuracy rises sharply from 37.46\% to 47.16\% at Pass@1, with gains persisting through Pass@10 (59.56 → 67.11). GPT paraphrased inputs also show notable improvements.

Overall, the Lean 4 results reinforce the trend observed in Isabelle: paraphrasing can meaningfully shift model performance, and robustness to linguistic variation remains an open challenge for current autoformalization systems.

\begin{table*}[t]
\centering
\caption{Pass@K Evaluation for Autoformalization on ProofNet(Lean 4) and MiniF2F(Isabelle/HOL).
We report both Pass@K BLEU Accuracy (semantic fidelity) and Pass@K Compilation Accuracy (syntactic validity)
across varying $K$ values. Each section corresponds to the LLM used for formalization.}
\label{tab:passk_combined}
\begin{tabular}{lcccccccc}
\toprule
\multirow{2}{*}{\textbf{Model / Setting}} &
\multicolumn{4}{c}{\textbf{Pass@K BLEU Accuracy (\%)}} &
\multicolumn{4}{c}{\textbf{Pass@K Compilation Accuracy (\%)}} \\
\cmidrule(lr){2-5} \cmidrule(lr){6-9}
& K=1 & K=2 & K=5 & K=10 & K=1 & K=2 & K=5 & K=10 \\
\midrule
\multicolumn{9}{c}{\textbf{GPT-4o-mini (Formalization Model)}} \\
\midrule
Ground Truth NL→Isabelle           & 4.84 & 5.57 & 6.54 & 7.41 & 12.30 & 20.08 & 31.56 & 36.07 \\
GPT Paraphrased NL→Isabelle        & 8.31 & 9.39 & 10.92 & 11.87 & 20.90 & 25.41 & 34.84 & 37.30 \\
Claude Paraphrased NL→Isabelle     & 4.77 & 5.36 & 6.32 & 7.00 & 13.93 & 20.90 & 34.02 & 39.75 \\
\midrule
Ground Truth NL→Lean 4               & 10.12 & 10.78 & 11.50 & 12.21 & 9.70 & 13.74 & 17.52 & 19.67 \\
GPT Paraphrased NL→Lean 4            & 15.10 & 16.00 & 16.97 & 17.75 & 7.27 & 11.05 & 15.63 & 18.32 \\
Claude Paraphrased NL→Lean 4         & 15.36 & 16.06 & 17.50 & 18.44 & 7.00 & 9.97 & 14.55 & 16.44 \\
\midrule
\multicolumn{9}{c}{\textbf{Claude-3.7-Sonnet (Formalization Model)}} \\
\midrule
NL→Isabelle                        & 6.35 & 7.30 & 9.06 & 10.29 & 10.66 & 19.26 & 46.31 & 57.79 \\
GPT Paraphrased NL→Isabelle        & 13.90 & 15.24 & 18.73 & 20.25 & 12.70 & 20.49 & 47.13 & 53.28 \\
Claude Paraphrased NL→Isabelle     & 10.25 & 12.12 & 14.56 & 15.68 & 8.20 & 16.39 & 54.92 & 70.90 \\
\midrule
NL→Lean 4                            & 20.58 & 23.37 & 26.51 & 27.83 & 37.46 & 46.09 & 55.52 & 59.56 \\
GPT Paraphrased NL→Lean 4            & 26.98 & 28.76 & 31.63 & 32.70 & 45.28 & 56.60 & 65.22 & 69.00 \\
Claude Paraphrased NL→Lean 4         & 29.25 & 31.90 & 34.46 & 35.78 & 47.16 & 54.98 & 61.99 & 67.11 \\
\bottomrule
\end{tabular}
\end{table*}

\section{Conclusion}
Our results demonstrate that even when paraphrased statements maintain a high semantic similarity to their ground truth counterparts, current LLM-based autoformalization systems exhibit sensitivity in both semantic fidelity and syntactic validity. These results align with and extend the generalizability of recent findings in the text-to-SQL domain \cite{oracle-sql}, which similarly report that LLMs are highly sensitive to minor linguistic perturbations while preserving semantic meaning. Thus, there is a need for more robust autofomalization pipelines that can mitigate this sensitivity to minor linguistic perturbations, ensuring more consistent results.

\section{Limitations}
This paper performs a cross-evaluation of paraphrased NL statements for two models, future research should expand this to more models to see if the sensitivity to paraphrased inputs persists, or if some models are more robust. Additionally, we perform an evaluation on semantically similar paraphrased NL statements but this work does not explore the results for low semantic similarity, ambiguous, or inconsistent proof requests. Finally, this work does not perform a systematic evaluation of the semantic or compilation error categories, which could provide deeper insights into the sensitivity of paraphrased inputs.

\textit{}

\bibliography{aaai2026}

@article{oracle-sql,
  author    = {Mohammadtaher Safarzadeh and Afshin Oroojlooyjadid and Dan Roth},
  title     = {Evaluating {NL2SQL} via {SQL2NL}},
  journal   = {arXiv preprint arXiv:2509.04657v1},
  year      = {2025},
  url       = {https://arxiv.org/abs/2509.04657},
  note      = {Affiliation: Oracle AI}
}

@article{minif2f,
  title        = {MiniF2F: a cross‐system benchmark for formal Olympiad‐level mathematics},
  author       = {Kunhao Zheng and Jesse Michael Han and Stanislas Polu},
  journal      = {arXiv preprint arXiv:2109.00110},
  year         = {2021},
  url          = {https://arxiv.org/abs/2109.00110}
}

@article{papineni2002bleu,
  title     = {BLEU: a method for automatic evaluation of machine translation},
  author    = {Papineni, Kishore and Roukos, Salim and Ward, Todd and Zhu, Wei-Jing},
  journal   = {Proceedings of the 40th Annual Meeting of the Association for Computational Linguistics (ACL)},
  pages     = {311--318},
  year      = {2002},
  url       = {https://aclanthology.org/P02-1040}
}

@article{proof,
  title        = {ProofNet: Autoformalizing and Formally Proving Undergraduate-Level Mathematics},
  author       = {Azerbayev, Zhangir and Piotrowski, Bartosz and Schoelkopf, Hailey and Ayers, Edward W. and Radev, Dragomir and Avigad, Jeremy},
  journal      = {arXiv preprint arXiv:2302.12433},
  year         = {2023},
  url          = {https://arxiv.org/abs/2302.12433}
}

@inproceedings{lean,
  title        = {The Lean Theorem Prover (System Description)},
  author       = {de Moura, Leonardo and Kong, Soonho and Avigad, Jeremy and Van Doorn, Floris and von Raumer, Jakob},
  booktitle    = {Automated Deduction – CADE-25},
  series       = {Lecture Notes in Computer Science},
  volume       = {9195},
  pages        = {378--388},
  year         = {2015},
  publisher    = {Springer},
  doi          = {10.1007/978-3-319-21401-6_26},
  url          = {https://doi.org/10.1007/978-3-319-21401-6_26}
}

@inproceedings{isabelle,
  author    = {Tobias Nipkow and Lawrence C. Paulson and Markus Wenzel},
  title     = {Isabelle/HOL — A Proof Assistant for Higher-Order Logic},
  booktitle = {Theorem Proving in Higher Order Logics (TPHOLs 2002)},
  series    = {Lecture Notes in Computer Science},
  volume    = {2283},
  publisher = {Springer},
  year      = {2002},
  pages     = {1--16},
  doi       = {10.1007/3-540-45949-9_1},
  url       = {https://doi.org/10.1007/3-540-45949-9_1}
}

@misc{gpt,
  title        = {{GPT-4o mini}: Advancing Cost-Efficient Intelligence},
  author       = {{OpenAI}},
  year         = {2024},
  howpublished = {\url{https://openai.com/index/gpt-4o-mini-advancing-cost-efficient-intelligence/}},
  note         = {Accessed: 2025-10-31}
}

@misc{claude,
  title        = {{Claude 3.7 Sonnet}: Hybrid Reasoning Model from Anthropic},
  author       = {{Anthropic}},
  year         = {2025},
  howpublished = {\url{https://www.anthropic.com/news/claude-3-7-sonnet}},
  note         = {Accessed: 2025-10-31}
}

@misc{LLM,
      title={Autoformalization with Large Language Models}, 
      author={Yuhuai Wu and Albert Q. Jiang and Wenda Li and Markus N. Rabe and Charles Staats and Mateja Jamnik and Christian Szegedy},
      year={2022},
      eprint={2205.12615},
      archivePrefix={arXiv},
      primaryClass={cs.LG},
      url={https://arxiv.org/abs/2205.12615}, 
}

@misc{thor,
      title={Thor: Wielding Hammers to Integrate Language Models and Automated Theorem Provers}, 
      author={Albert Q. Jiang and Wenda Li and Szymon Tworkowski and Konrad Czechowski and Tomasz Odrzygóźdź and Piotr Miłoś and Yuhuai Wu and Mateja Jamnik},
      year={2022},
      eprint={2205.10893},
      archivePrefix={arXiv},
      primaryClass={cs.AI},
      url={https://arxiv.org/abs/2205.10893}, 
}

@misc{leandojo,
      title={LeanDojo: Theorem Proving with Retrieval-Augmented Language Models}, 
      author={Kaiyu Yang and Aidan M. Swope and Alex Gu and Rahul Chalamala and Peiyang Song and Shixing Yu and Saad Godil and Ryan Prenger and Anima Anandkumar},
      year={2023},
      eprint={2306.15626},
      archivePrefix={arXiv},
      primaryClass={cs.LG},
      url={https://arxiv.org/abs/2306.15626}, 
}

@book{lex,
  title        = {Measuring Lexical Diversity},
  author       = {Tweedie, Fiona J. and Baayen, R. Harald},
  year         = {1998},
  publisher    = {Springer},
  booktitle    = {Computers and the Humanities},
  volume       = {32},
  number       = {5},
  pages        = {323--352},
  doi          = {10.1023/A:1002199404110}
}

@article{lean4,
      title={DeepSeek-Prover-V1.5: Harnessing Proof Assistant Feedback for Reinforcement Learning and Monte-Carlo Tree Search}, 
      author={Huajian Xin and Z. Z. Ren and Junxiao Song and Zhihong Shao and Wanjia Zhao and Haocheng Wang and Bo Liu and Liyue Zhang and Xuan Lu and Qiushi Du and Wenjun Gao and Qihao Zhu and Dejian Yang and Zhibin Gou and Z. F. Wu and Fuli Luo and Chong Ruan},
      year={2024},
      eprint={2408.08152},
      archivePrefix={arXiv},
      primaryClass={cs.CL},
      url={https://arxiv.org/abs/2408.08152}, 
}

@inproceedings{sbert,
  title        = {Sentence-BERT: Sentence Embeddings Using Siamese BERT-Networks},
  author       = {Reimers, Nils and Gurevych, Iryna},
  booktitle    = {Proceedings of the 2019 Conference on Empirical Methods in Natural Language Processing (EMNLP) and the 9th International Joint Conference on Natural Language Processing (IJCNLP)},
  pages        = {3982--3992},
  year         = {2019},
  organization = {Association for Computational Linguistics},
  doi          = {10.18653/v1/D19-1410},
  url          = {https://aclanthology.org/D19-1410/}
}


\appendix
\section{Appendix}
\label{sec:appendix}

\subsection{Prompts for Autoformalization Evaluation}
\label{sec:prompts}

We provide the exact prompts used for our autoformalization experiments. 
These prompts were used with \texttt{GPT-4o-mini/Claude-3.7-sonnet} for both directions:
(1) Natural Language to Isabelle/ Lean 4 Formalization (NL→Formal), and 
(2) Paraphrased evaluation (Formal→NL→Formal). 
All generations were executed with temperature = 0.0, top\_p = 1.0, max tokens = 500 (for single generation), max tokens = 5000 (for multiple generations), no context between tasks in roundtrip.

\subsubsection{Prompt 1: Ground-Truth NL → Formal (Isabelle)}
\begin{quote}
\small
\textbf{System:} You are a helpful assistant that translates between formal logic and natural language.

\textbf{User:}
Translate the following Natural Language statement into a \texttt{[Isabelle]} Formal Statement that conveys the exact same logical meaning.
Generate only the \texttt{[Isabelle]} Formal Statement, without any additional commentary or explanation.

Natural Language Statement:
\texttt{[natural\_s]}

Formal Statement:

\end{quote}

\subsubsection{Prompt 2: Ground-Truth Formal → NL (Isabelle/Lean 4)}
\begin{quote}
\small
\textbf{System:} You are a helpful assistant that translates between formal logic and natural language.

\textbf{User:} Translate the following \texttt{[Isabelle/Lean 4]} statement into a clear natural language question that conveys the exact same logical meaning. Generate only the natural language question written in LaTex, without any additional commentary or explanation.

Formal Statement:
\texttt{[formal\_s]}

Natural Language Statement:
\end{quote}

\subsubsection{Prompt 3: NL → Formal, Multiple Generation (Isabelle)}
\begin{quote}
\small
\textbf{System:} You are a helpful assistant that translates between formal logic and natural language.

\textbf{User:}
Translate the following natural language statement into 10 clear \texttt{[Isabelle]} statements that conveys the exact same logical meaning. Generate a numbered list of 10 unique \texttt{[Isabelle]} statements, without any additional commentary or explanation.

Natural Language Statement:
\texttt{[natural\_s]}

Formal Statement:
1.

\end{quote}

\subsubsection{Prompt 4: Ground-Truth NL → Formal (Lean 4)}
\begin{quote}
\small
\textbf{System:} You are a helpful assistant that translates between formal logic and natural language.

\textbf{User:} 
Translate the following Natural Language (or LaTeX) statement into a clear, valid \texttt{[Lean 4]} theorem that convey the same logical meaning without any additional commentary or explanation.
Requirements:
        \begin{enumerate}
        \item Output (Translated \texttt{[Lean 4]} statement) must be a string  assigned to the output field message.
        \item Translated statement must:
        \begin{itemize}
            \item  Begin with 'theorem'
            \item Be a self-contained \texttt{[Lean 4]} statement
            \item Do not include import lines
            \item Not include 'by', 'sorry', or 'import'
            \item Encode any necessary assumptions in variable names or hypotheses
        \end{itemize}
        \item Make reasonable assumptions if the natural language statement is underspecified.
        \item Do not add any commentary or explanations.
        
        \end{enumerate} 
\texttt{[natural\_s]}
\end{quote}
\subsubsection{Prompt 5: NL → Formal, Multiple Generation (Lean 4)}
\begin{quote}
\small
\textbf{System:} You are a helpful assistant that translates between formal logic and natural language.

\textbf{User:}
Translate the following Natural Language (or LaTeX) statement into  exactly 10 clear, valid \texttt{[Lean 4]} theorems that convey the same logical meaning without any additional commentary or explanation.
Requirements:
        \begin{enumerate}
        \item Output must be a Python list assigned to the output field message.
        \item The list must contain exactly 10 string elements (each element is a valid \texttt{[Lean 4]} translation), no more, no less.
        \item Each element must:
        \begin{itemize}
            \item  Begin with 'theorem'
            \item Be a single, self-contained \texttt{[Lean 4]} statement
            \item Do not include import lines
            \item Not include 'by', 'sorry', or 'import'
            \item Encode any necessary assumptions in variable names or hypotheses
        \end{itemize}
        \item Make reasonable assumptions if the natural language statement is underspecified.
        \item Do not add any commentary or explanations.
        \end{enumerate} 
\texttt{[[natural\_s]}
\end{quote}

\end{document}